\documentclass[conference]{IEEEtran}

\usepackage{graphicx}%
\usepackage{multirow}%
\usepackage{amsmath,amssymb,amsfonts}%
\usepackage{amsthm}%
\usepackage{mathrsfs}%
\usepackage{xcolor}%
\usepackage{textcomp}%
\usepackage{manyfoot}%
\usepackage{booktabs}%
\usepackage{algorithm}%
\usepackage{algorithmicx}%
\usepackage{algpseudocode}%
\usepackage{listings}%
\usepackage{subcaption}
\usepackage{tabularx}
\newcolumntype{Y}{>{\raggedright\arraybackslash}X}

\usepackage{natbib}
\usepackage{physics}

\newcommand{\CF}{\Lambda^\mu_d(\mathbf{x})}

\newtheorem{theorem}{Theorem}[subsection]
\IEEEoverridecommandlockouts

\begin{document}


\title{Leveraging the Christoffel Function for Outlier Detection in Data Streams}

\author{
\IEEEauthorblockN{
K\'evin Ducharlet\IEEEauthorrefmark{1}\IEEEauthorrefmark{2},
Louise Trav\'e-Massuy\`es\IEEEauthorrefmark{1},
Jean-Bernard Lasserre\IEEEauthorrefmark{1},
Marie-V\'eronique Le Lann\IEEEauthorrefmark{1},
Youssef Miloudi\IEEEauthorrefmark{2}
}
\IEEEauthorblockA{\IEEEauthorrefmark{1}LAAS-CNRS, ANITI, University of Toulouse, CNRS, INSA, Toulouse, France\\
Emails: kevin.ducharlet@berger-levrault.com, louise@laas.fr, lasserre@laas.fr, mvlelann@laas.fr}
\IEEEauthorblockA{\IEEEauthorrefmark{2}Carl Berger-Levrault, Limonest, France\\
Email: youssef.miloudi@berger-levrault.com}
}
\maketitle

\begin{abstract}
Outlier detection holds significant importance in the realm of data mining, particularly with the growing pervasiveness of data acquisition methods. The ability to identify outliers in data streams is essential for maintaining data quality and detecting faults. However, dealing with data streams presents challenges due to the non-stationary nature of distributions and the ever-increasing data volume. While numerous methods have been proposed to tackle this challenge, a common drawback is the lack of straightforward parameterization in many of them. This article introduces two novel methods: DyCF and DyCG. DyCF leverages the Christoffel function from the theory of approximation and orthogonal polynomials. Conversely, DyCG capitalizes on the growth properties of the Christoffel function, eliminating the need for tuning parameters. Both approaches are firmly rooted in a well-defined algebraic framework, meeting crucial demands for data stream processing, with a specific focus on addressing low-dimensional aspects and maintaining data history without memory cost. A comprehensive comparison between DyCF, DyCG, and state-of-the-art methods is presented, using both synthetic and real industrial data streams. The results show that DyCF outperforms fine-tuning methods, offering superior performance in terms of execution time and memory usage. DyCG performs less well, but has the considerable advantage of requiring no tuning at all.
\end{abstract}



\section{Introduction}\label{sec:intro}

The identification and examination of uncommon observations play a crucial role in data mining, as they may signal data corruption or faulty behavior. Such unusual observations can be categorized as outliers, anomalies, out-of-distribution samples, or novelties. We specifically adopt the term "outlier" along with Hawkins' definition \citep{hawkins1980identification} of ``an observation which deviates so much from the other observations as to arouse suspicions that it was generated by a different mechanism''. Outliers carry valuable information about underlying processes, making them especially relevant in applications like network traffic analysis \citep{zhang2006anomaly}, medical diagnosis \citep{dreiseitl2010outlier}, and fraud detection \citep{malini2017analysis}, where detecting abnormal behavior is crucial. Furthermore, outliers can significantly disrupt various machine learning methods employed in tasks such as prediction or decision-making, necessitating their removal to ensure the accuracy of the results obtained.

In the contemporary landscape, various data sources such as wireless sensor networks, social networks, medical systems, web traffic, and online transactions continuously generate data. The resulting datasets exhibit characteristics of uncertainty and continuous evolution, posing significant challenges for outlier detection in this dynamic environment. Traditional methods designed for \textit{batch datasets} typically seek a mapping function that assigns an outlierness score to new samples based on the observation of an entire set of historical samples. In some cases, only these historical samples can receive an outlierness score, leaving new, unseen samples unassessed. In other scenarios, methods may categorize new samples as inliers or outliers, but the mapping function remains static and does not adapt over time. However, for effective outlier detection in data streams, methods must actively seek a mapping function that adjusts to new samples and grapple with data streams of infinite length.

In the context of outlier detection, labels are frequently unavailable~\citep{goix2016how}, making it uncertain whether historical samples are genuinely outliers. While batch scenarios allow for preprocessing to label data and satisfy supervised learning conditions, obtaining a reliable set of normal samples or choosing known outliers can facilitate semi-supervised tasks. However, in the realm of data streams, the continuous evolution of data distribution renders labeling impractical and it can swiftly become outdated. Consequently, outlier detection methods must operate in an unsupervised manner. The absence of labels also introduces challenges in fine-tuning these methods, as evaluating their performance becomes arduous without labeled data.

This paper focuses on unsupervised outlier detection for low-dimensional data streams. We highlight the applicability of the Christoffel function (CF), a well-established concept in the theory of approximation and orthogonal polynomials, in addressing this challenge. Our contributions encompass (1) adapting the CF to assess outliers in data streams, resulting in the \emph{Dynamic Christoffel Function} (DyCF) method, (2) introducing a tuning-free approach called \emph{Dynamic Christoffel Growth} (DyCG), capitalizing on the asymptotic growth properties of the CF, and (3) conducting comparisons with several state-of-the-art methods using synthetic and real industrial data streams. 

The structure of this paper is as follows. In Section~\ref{sec:related}, we offer an overview of outlier detection in data streams, delving into the current state of the art. Section~\ref{sec:christoffel} introduces the Christoffel function (CF), illustrating its ability to effectively capture the support of a theoretical measure from a set of samples. Additionally, we compare it with the closely related method, Kernel Density Estimation (KDE). In Section~\ref{sec:datastreams}, we present DyCF, an adaptation of the CF for handling data streams along with its tuning-free enhancement, DyCG. Section~\ref{sec:eval} provides the results of DyCF and DyCG, comparing them with state-of-the-art methods. Finally, Section~\ref{sec:conclu} concludes the paper by discussing the results and suggesting potential enhancements for DyCF and DyCG, outlining avenues for future research.


\section{Problem formulation}\label{sec:problem}

The problem that we consider is embedded unsupervised outlier detection in low dimensional data streams issued from low capacity sensoring devices. 

The peculiarities of data streams that require consideration include \cite{sadik2014research}:
\begin{itemize}
    \item \emph{Transiency}: the significance of each data point diminishes over time; therefore, it should be processed promptly upon measurement.
    \item \emph{Time dependency}: each data point is linked to a timestamp, which must be taken into account either as an attribute or in the order of arrival. In both scenarios, a data point is assessed in comparison to other points within the same temporal context.
    \item \emph{Infinity}: as measurements are continuously generated, data streams constitute theoretically infinite sequences of samples and, therefore, cannot be stored in memory entirely, particularly in low memory sensors. Thus methods should opt for a summary of the dataset rather than attempting to store the entire sequence.
    \item \emph{Arrival rate}: the arrival rate may vary over time, but it is imperative to process points immediately upon measurement. Therefore, the algorithm's execution time must be sufficiently brief. In the event of a variable arrival rate, it might be necessary to adapt the process and be willing to compromise on accuracy.
    \item \emph{Concept drift}: in many cases, data distribution is non-stationary\footnote{Non-stationary distributions have means, variances, and covariances that change over time. Non-stationary behaviors can be trends, cycles, random walks, or combinations of the three.} making outlier detection methods that assume a fixed distribution unsuitable.
    \item \emph{Uncertainty}: In various application scenarios, measurements can be influenced by environmental disturbances. This justifies the use of outlier detection methods.
    \item \emph{Multi-dimensionality}: while not exclusive to data streams, some challenges are associated with high dimensionality. In our work, we concentrate on problems that are low-dimensional yet multi-dimensional.
    \item \emph{Embeddedness}: an additional consideration is related to the concept of edge computing. In various samples, especially within wireless sensor networks, computing capabilities are integrated into objects with limited capacities, including memory and CPU.
\end{itemize}

In this context, there is a demand for approaches that exhibit the following characteristics, such as DyCF and DyCG proposed in this paper : 
\begin{itemize}
    \item frugality allowing to embed outlier detection models in devices, 
    \item fast update to match incoming measurement frequency,
    \item little or no fine-tuning to meet automation and generalization needs,
    \item explainability and interpretability so that human operators understand the results easily.
\end{itemize}
This being said, these properties exclude deep learning methods.


\section{Related Work}\label{sec:related}

Outlier detection has been a research subject for a long time in different communities, starting with statisticians and the works of Edgeworth in the end of the 19\textsuperscript{th} century \citep{edgeworth1887xli}. With more than a century of interest in outlier detection, a lot of different methods have been proposed and a significant number of surveys tackle the task of listing, describing, categorizing and comparing these methods, e.g., \citep{wang2019progress,chandola2009anomaly,ben_gal2005outlier,ruff2021unifying}. 

Depending on the context, outlier detection methods are usually separated into three groups: 1) supervised models that rely on the availability of datasets labeled with the outlierness status of samples, 2) semi-supervised methods that rely on datasets in which only normal samples are labeled, 3) unsupervised methods that can accept datasets without any information on outlierness. Unsupervised methods are recognized to be less precise than supervised methods due to the absence of information. However, as mentioned earlier, a limitation of supervised methods in the case of data streams is the potential obsolescence of labels resulting from distribution changes. Consequently, unsupervised methods become the sole option when dealing with data streams. For this reason, extensive research has been conducted on outlier detection methods for data streams. The reader can refer to surveys that concentrate on specific techniques \citep{tran2016distance, thakkar2016survey, salehi2018survey}, or those that survey the advancements of the field \citep{zhang2013advancements,sreevidya2014survey,wang2019progress}.

Initially considered, it seems interesting to adapt time series methods \citep{duraj2021outlier}, for example ARIMA models \cite{asteriou2011arima}, prediction models based on exponential smoothing \citep{hyndman2002state} and LSTM (Long Short-Term Memory) \citep{malhotra2015long}. These methods employ trends and seasonal patterns to forecast future data points from past observations. Anomaly detection can then be based on comparing forecasted points to actual measurements. However, these methods are not suitable for data streams because the learned model fails to evolve with new incoming measurements. While trend and seasonality can bring about alterations in the distribution, these changes must follow a regular pattern for models to make accurate predictions, and this regularity is not guaranteed in the context of data streams.

The three main families of outlier detection methods for data streams are methods based on dynamic clustering, those relying on nearest neighbors (kNN) logic, and statistical methods. A common strategy for making the two latter methods applicable to data streams involves the utilization of windowing techniques. Data windows retain a constant number of points, capturing the current temporal context and distribution. This effectively addresses the necessities for \emph{transiency}, \emph{infinity}, and \emph{concept drift}. Four windowing techniques are known \citep{salehi2018survey}:
\begin{itemize}
    \item \emph{Landmark windows} set a point as a landmark and process data between this point and the current data point.
    \item \emph{Sliding windows} process the last $W$ data points, $W$ being the size of the window.
    \item \emph{Damped windows} consider all the points but each point is assigned a diminishing weight corresponding to its age.
    \item \emph{Adaptive windows} are like sliding windows but their size varies with the speed evolution of points; the faster the distribution changes, the smaller the window.
\end{itemize}

Note that simply combining static methods with windowing techniques often proves inefficient. Many methods encounter challenges when dealing with swift model updates because they often require large window sizes to achieve satisfying results. This goes hand in hand with the fact that they are not engineered to be updated, necessitating the computation of a new model for each subsequent window, a process that can be time-consuming. 

\paragraph{Dynamic clustering}
Clustering methods group samples in space according to some similarity criterion and have been used to detect outliers based on one of the following assumptions~\citep{chandola2009anomaly}:
\begin{itemize}
    \item ``normal samples belong to clusters while outliers do not'' in the case where the method includes a rejection mechanism,
    \item ``normal samples are close to their closest centroïd (center of cluster) while outliers are far'' in the case where the method assigns all samples to clusters,
    \item ``dense clusters are normal and sparse clusters are outlying''.
\end{itemize}
To adapt to data streams, dynamic clustering methods make statistical properties of clusters or micro-clusters to evolve through time \citep{zhang1996birch, aggarwal2003framework, roa2019dyclee}. Their main advantage is that they tackle the \emph{notion of infinity} since it is not necessary to keep all the dataset in memory. However, they are often criticized because they have not been developed for outlier detection purposes but mainly for clustering \citep{thakkar2016survey}.

\paragraph{Methods relying on kNN}
Many methods for data streams consider outliers through the $k$ nearest neighbors (kNN) principle. These methods can be divided into two groups:
\begin{itemize}
    \item Methods for detecting outliers define an outlier as a sample with at most a proportion $r$ of points within a certain distance $D$, which can be thought of as having at most $k$ neighbors within a distance $d$ or being no farther than $d$ from the $k$-th nearest neighbor \citep{knorr1998algorithms}. These methods employ windowing techniques to reduce the number of samples stored in memory and use specialized data structures for efficient addition, removal, and kNN searches. Among these methods, a study by \cite{tran2016distance} finds that MCOD \citep{kontaki2011continuous} is the most efficient, although it has a limitation related to window size dependency.
    \item Methods adapting the well-known LocalOutlierFactor (LOF) algorithm \citep{breunig2000lof}. The LOF is a measure of how local density of a sample compares to local density of its neighbors. On the addition of a new sample, the incremental LOF (iLOF) uses the fact that only a fixed number of samples need to be updated to reduce the computational complexity \citep{pokrajac2007incremental}. However, the required search for kNN and reversed kNN remains costly, which explains that several methods have proposed to approximate the LOF measure \citep{karimian2012i, salehi2016fast, na2018dilof, huang2020tadilof}.
\end{itemize}

\paragraph{Statistical methods}
Statistical methods make the hypothesis that data samples are generated by a statistical distribution and that outliers belong to areas of low probability \citep{zhang2013advancements}. 

Parametric methods are unsuitable to non stationary distributions for they make the hypothesis of a predefined distribution and estimate its parameters. However, the method SmartSifter~\citep{yamanishi2004line} is worth mentioning as a statistical method offering both parametric and non-parametric solutions, showing better results with its parametric version. Its main advantage is that it is able to deal with categorical and continuous variables. The parametric version of SmartSifter uses Gaussian Mixture Models (GMM).

On the non parametric side, histogram construction is a candidate  in univariate settings. The number of elements falling in a cell of the histogram reflects the probability of a sample falling into this cell. An evident benefit is the ease with which new data points can be seamlessly incorporated into the model. Quantile sketches are also worth mentioning as an optimal solution for the resolution of this problem in the context of data streams~\citep{karnin2016optimal, zhao2021kll}. Interestingly, quantile sketches can be approximated based on moments~\citep{gan2018moment}, which, as we will see, are also at the heart of the proposed methods. 

To address multivariate scenarios, it's a common practice to construct individual histograms for each variable and subsequently compute a score by aggregating the scores from these separate histograms, a technique employed by HBOS (Histogram-based Outlier Score) as described in Goldstein's work \citep{goldstein2012histogram}. However, this approach encounters limitations in high-dimensional contexts, as it fails to consider the interdependencies between variables.

A more advanced solution is given by \emph{Kernel Density Estimation (KDE)} methods, also known as Parzen-Rosenblatt methods \citep{parzen1962estimation}. KDE (Kernel Density Estimation) shares similarities with histogram construction but incorporates a concept of continuity, offering an approximation of the probability density function (pdf). 
In the univariate case \citep{parzen1962estimation}, the estimator of the density function $f$ of $n$ samples $\mathcal{X}=\{\mathbf{x}_i,i=1,\dots,n\}$ issued from the theoretical measure $\mu$ is $\tilde{f_h}(x)=\frac{1}{n}\sum_{i=1}^{n}{\mathbf{K}_h(\mathbf{x}-\mathbf{x}_i)}$,
where $\mathbf{K}_h(u)=\frac{1}{h}\mathbf{K}(\frac{u}{h})$, $\mathbf{K}$ being the kernel function \footnote{The kernel function is often chosen as Gaussian or as the Epanechnikov one.} and $h$ being the bandwidth parameter that affects the influence area of each sample, or in other words, the smoothness of the function. Multivariate KDE (KDE) uses multivariate kernels $\mathbf{K_H}(u)=\abs{\mathbf{H}}^{-1/2}\mathbf{K}(\mathbf{H}^{-1/2}u)$, where $\mathbf{H}$ is a symmetric positive definite $p\times p$ bandwidth matrix \citep[§2.3.1]{langren_e2019fast}.
KDE methods give a better approximation than histograms and are able to deal with multivariate cases although their complexity raises quickly with the amount of variables. The KDE based method proposed in \citep{kristan2011multivariate} applies to data streams, as well as the non-parametric version of SmartSifter \citep{yamanishi2004line}.

\paragraph{Contributions of the proposed CF based methods}

The methods that we propose in this paper, namely DyCF and DyCG, can be positioned as statistical methods. The CF indeed captures the statistics of the dataset. Among the methods discussed in this section, KDE methods are undoubtedly the most closely related. However, the CF introduces a distinct perspective compared to KDE, as it identifies the theoretical probability measure of a set of samples using the statistical moments. 

DyCF and DyCG advance the state of the art and bring contributions in three directions:
\begin{itemize}
    \item they are based on solid theoretical foundations as they inherit the proven properties of the CF, 
    \item they satisfy all data stream requirements, in particular they achieve fast model update on the arrival of new samples while retaining memory of past data,
    \item they require very little tuning, i.e. only one hyperparameter for DyCF, or no tuning at all for DyCG, hence avoiding the painful and tedious tuning phase required by the state of the art methods.
\end{itemize}


\section{The Christoffel function for outlier detection}\label{sec:christoffel}

Prior to this section, we provide, in Table~\ref{tab:math_not}, a list of mathematical notations used for characterizing the Christoffel function.

\begin{table}[htbp]
    \centering
    \begin{tabular}{ll}
        \toprule
        Notation & Description \\
        \midrule
        $\mu$ & A measure with support $\Omega\subset\mathbb{R}^p$ \\
        $\Omega$ & Support of $\mu$ \\
        $p$ & Dimension of the support $\Omega$ \\
        $d$ & Parameter of the Christoffel function \\
        $\Lambda^\mu_d$ & Christoffel function with degree $d$ \\
        $Q_{\mu,d}$ & Scoring function based on $\Lambda^\mu_d$ \\
        $\Omega_\gamma$ & Level set $\Omega_\gamma:=\{\mathbf{x}: \CF^{-1}\leq \gamma\}$ \\
        $\gamma_{d,p}$ & Define a level set $\Omega_{\gamma_{d,p}}$ with $\gamma_{d,p} = d^{3p/2}$ \\
        $\mathbf{v}_d(X)$ & Vector of monomials of degree less than $d$ \\
        $s_p(d)$ & Size of $\mathbf{v}_d(X)$, equals to $\binom{p+d}{d}$ \\
        $y_\alpha(\mu)$ & Moment $\alpha$ of $\mu$ \\
        $M_d(\mu)$ & Matrix of moments of size $s_p(d)\times s_p(d)$ \\
        $\mathcal{X}$ & Set of $n$ observations from $\mu$ \\
        $\mu_n$ & The empirical measure supported by $\mathcal{X}$ \\
        \bottomrule
    \end{tabular}
    \caption{Table of notations}
    \label{tab:math_not}
\end{table}

\subsection{Main properties of the Christoffel function}

The Christoffel-Darboux Kernel (CD-Kernel) and the associated Christoffel function (CF) are well-known tools from the theory of approximation and orthogonal polynomials. Although they have been largely ignored in analysis of discrete data, recent results show that some peculiar properties of the CF can be valuable \citep{lasserre2019empirical,lasserre2022christoffel}.

The CD-Kernel and the CF are associated with a measure $\mu$ with support $\Omega\subset \mathbb{R}^p$, usually compact with nonempty interior, empirically represented by the set of available $p$-variate points. They also depend on a parameter $d$ defining the highest degree of monomials that index the moment matrix of the measure $\mu$ and is involved in the definition of the CF. 

The CF is hence denoted $\Lambda^\mu_d$, parameterized by the measure $\mu$ and by the degree $d$. One of its main and salient features is its ability to encode the support $\Omega$. In particular, for dimensions $p=2$ or $p=3$, one observes that the level set 
$\Omega_\gamma:=\{\mathbf{x}: \CF^{-1}\leq \gamma\}$, defined for some $\gamma \in \mathbb{R}_+$, captures the geometric shape of $\Omega$ quite accurately, even for low degrees $d$. Used as a tuning parameter, \emph{the integer $d$ gives a trade off between regularity (with small values of $d$) and fitness (with higher values) of the shape}. 

As presented formally in section \ref{sec:formal_def} and given a measure $\mu$, the associated CF is obtained from the moment matrix of $\mu$. Now, moments serve to quantify three essential parameters of distributions: location, shape and scale. The location of a distribution pertains to the position of its center of mass. Scale, on the other hand, denotes the extent to which a distribution is spread out, with the scale factor influencing the stretching or compression of the distribution. Lastly, the shape of a distribution encompasses its overall geometry, including characteristics such as bimodality, asymmetry, and heavy-tailedness. Consequently, the first moment delineates a distribution's location, the second moment characterizes its scale, and higher moments collectively elucidate its shape. The CF inherits this knowledge through the moment matrix, which intuitively explains why it can be a powerful tool for data analysis.

Previous works~\citep{lasserre2019empirical, lasserre2022christoffel} have shown how some of the CF's key properties can be helpful to address important problems like density approximation, support inference and outlier detection, where the measure of interest is now a \emph{discrete measure} $\mu_n$ whose support is a finite set (or ``cloud") of $n$ data points (or samples) sampled from $\mu$.

When going from $\mu$ on $\Omega$ to the empirical measure $\mu_n$ on the data set of $n$ samples, it is important to relate $n$ and $d$ so that the empirical Christoffel function $\Lambda^{\mu_n}_d$ captures essential features of the population. For fixed $d$, the fact that $\Lambda^{\mu_n}_d$ and $\Lambda^\mu_d$ share the same properties is essentially dictated by the \emph{Strong Law of Large Numbers}; see e.g. \citep{lasserre2022christoffel} (§6.2), and so it is sufficient that $n$ is large enough compared to $d$, which is often the case in practice for small $d$. When $d$ increases, the condition relating the sample size $n$ and the degree $d$ for $\Lambda^{\mu_n}_d$ to be close to $\Lambda^\mu_d$ is proven in \citep{lasserre2022christoffel} (§6.3). In \citep{vu2022rate} and \citep{lasserre2022christoffel} (Corollary 6.3.2), one can find a recipe to choose $n$ and $d$ in an appropriate manner.

On top of that, note that having $n$ large enough is not an issue regarding computational complexity since the empirical CF, as defined later in equations (\ref{eq:CF}-\ref{eq:empirical_M_d}), does not depend on the size of the dataset but solely on the number of variables $p$ and the degree $d$.

\subsection{Formal definition of the Christoffel function}\label{sec:formal_def}

Let $X=(X_1,X_2,...,X_p)\in\mathbb{R}^p$ and let $\alpha=(\alpha_i)_{i=1...p}\in\mathbb{N}^p$ be the vector of exponents (degrees) associated to each variable for the monomial $X^{\alpha}:=X_1^{\alpha_1}X_2^{\alpha_2}...X_p^{\alpha_p}$ of total degree $\sum_{i=1}^p\alpha_i$. Let $\mathbf{v}_d(X)$ be the vector of all monomials of degree less than or equal to $d$ in the \textit{graded lexicographic order}\footnote{Graded lexicographic order means: 1) ordered according to ascending monomial degree and then 2) using lexicographic order on variables considering $X_1=a$, $X_2=b$, etc.}. The size of the vector $\mathbf{v}_d(X)$, denoted $s_p(d)$, depends on $p$ and $d$ and is equal to $\binom{p+d}{d}$.

As defined in \citep{lasserre2022christoffel}, given a finite Borel probability measure $\mu$ on a compact set $\Omega\subset\mathbb{R}^p$ with nonempty interior, its moment matrix $M_d(\mu)$ is a real symmetric matrix with rows and columns indexed by the monomials of $\mathbf{v}_d(X)$. More precisely, letting \begin{equation}
    y_\alpha(\mu)\,:=\int_{\mathbb{R}^p}\mathbf{x}^\alpha d\mu(\mathbf{x})\label{eq:y_alpha}
\end{equation} be the moment $\alpha$ of $\mu$, this means that the element of the matrix, at row indexed by $\alpha=(\alpha_i)_{i=1...p}$ and column indexed by $\beta=(\beta_i)_{i=1...p}$, is $y_{\alpha+\beta}(\mu)=\int_{\mathbb{R}^p}\mathbf{x}^{\alpha+\beta} d\mu(\mathbf{x})$ with the notation $(\alpha+\beta)=(\alpha_i + \beta_i)_{i=1...p}$. For sample, in the case of $p=2$ and $d=2$ and denoting $y_\alpha=y_\alpha(\mu)$, the moment matrix is given by $$M_2(\mu):
\begin{matrix}
       & & 1      & X_1    & X_2    & X_1^2  & X_1X_2 & X_2^2  \\
       & &        &        &        &        &        &        \\
1      & & 1      & y_{1,0} & y_{0,1} & y_{2,0} & y_{1,1} & y_{0,2} \\
X_1    & & y_{1,0} & y_{2,0} & y_{1,1} & y_{3,0} & y_{2,1} & y_{1,2} \\
X_2    & & y_{0,1} & y_{1,1} & y_{0,2} & y_{2,1} & y_{1,2} & y_{0,3} \\
X_1^2  & & y_{2,0} & y_{3,0} & y_{2,1} & y_{4,0} & y_{3,1} & y_{2,2} \\
X_1X_2 & & y_{1,1} & y_{2,1} & y_{1,2} & y_{3,1} & y_{2,2} & y_{1,3} \\
X_2^2  & & y_{0,2} & y_{1,2} & y_{0,3} & y_{2,2} & y_{1,3} & y_{0,4} \\
\end{matrix}$$
$M_d(\mu)$ can also be written as \begin{equation}
    M_d(\mu)=\int_{\mathbb{R}^p}\mathbf{v}_d(\mathbf{x})^T\mathbf{v}_d(\mathbf{x}) d\mu(\mathbf{x}),\label{eq:M_d}
\end{equation} where the integral is understood elementwise. Note that $M_d(\mu)$ is positive definite for any $d$, i.e., $\mathbf{p}^T M_d(\mu)\mathbf{p} >0$
for every $0\neq\mathbf{p}\in\mathbb{R}^p$, and therefore $M_d(\mu)$ is non singular.

The CD-Kernel $K^\mu_d$ associated with $\mu$ is defined by
\begin{equation}\label{eq:Q}
(\mathbf{x},\mathbf{y})\mapsto K^\mu_d(\mathbf{x},\mathbf{y})\,:=\,\mathbf{v}_d(\mathbf{x})^T
M_d(\mu)^{-1}\mathbf{v}_d(\mathbf{y})\,,
\end{equation}
while the polynomial $Q_{\mu,d}$ reads
\begin{multline} \label{eq:inv_cf}
Q_{\mu,d}(\mathbf{x}):=K^\mu_d(\mathbf{x},\mathbf{x})=\mathbf{v}_d(\mathbf{x})^TM_d(\mu)^{-1}\mathbf{v}_d(\mathbf{x}),\,\\\quad\mathbf{x}\in\mathbb{R}^n\,.
\end{multline}

$Q_{\mu,d}$ is a sum-of-squares polynomial of degree $2d$ and the CF $\Lambda^\mu_d(\mathbf{x})$ is then defined by
\begin{equation}\label{eq:CF}
\mathbf{x}\mapsto \Lambda^\mu_d(\mathbf{x})^{-1}\,:=\,Q_{\mu, d}(\mathbf{x})\,,\quad \forall\mathbf{x}\in\mathbb{R}^n\,.
\end{equation}

\subsection{Outlier scoring with the Christoffel Function}\label{sec:cf_scoring}

In practical applications of outlier detection, only an \emph{empirical} moment matrix is available, associated with a discrete measure $\mu_n$ whose support is a set of $n$ observations $\mathcal{X}=\{\mathbf{x}_i,i=1,\dots,n\}$ sampled from a theoretical distribution $\mu$.
In this case, the empirical version of equations~(\ref{eq:y_alpha}) and~(\ref{eq:M_d}) respectively read 
\begin{equation}
    y_\alpha(\mu_n)=\frac{1}{n}\sum_{x\in\mathcal{X}}\mathbf{x}^\alpha\,,
\end{equation} and 
\begin{equation}
    M_d(\mu_n)=\frac{1}{n}\sum_{\mathbf{x}\in\mathcal{X}}\mathbf{v}_d(\mathbf{x})^T\mathbf{v}_d(\mathbf{x})\,.\label{eq:empirical_M_d}
\end{equation} 

Note that, considering $\mathcal{X}$ as a dataset, $M_d(\mu_n)$ can be seen as a summary or an encoding of this dataset. This property is very interesting because it avoids keeping in memory all the samples, which is definitively unacceptable when dealing with data streams (see the \textit{``Infinity''} peculiarity of data streams in Section \ref{sec:related}). 

Given a cloud of points $(\mathbf{x}_i)_{i\leq n}$ sampled from a theoretical measure $\mu$, the ability to capture the geometric shape of the support of the empirical measure $\mu_n$ comes with one valuable property of $\Lambda^\mu_d(\mathbf{x})^{-1}$. It has indeed been shown that, under some assumptions, the samples belonging to the support are confined by a specific level set $\Omega_{\gamma_{d,p}}$, where $\gamma_{d,p} = Cd^{3p/2}$ and $C$ is a problem-related constant \citep{lasserre2022christoffel}(Theorem 7.3.3). This level set will be used in the following sections, setting $C=1$. 

As a matter of fact, the level sets of $\Lambda^\mu_d(\mathbf{x})^{-1}$ match the density variations of the cloud of points $(\mathbf{x}_i)_{i\leq n}$, as shown in the illustrative example below, making of $\Lambda^\mu_d(\mathbf{x})^{-1}$ a good scoring function for outlier detection. 

Additionally, the model is contained in the moments matrix $M_d(\mu_n)$ of size $s_p(d)\times s_p(d)$, that does not depend on $n$, thereby fixing the memory size. The computational complexity is also limited since $\Lambda^\mu_d(\mathbf{x})^{-1}$ only requires computing $\mathbf{v}_d(\mathbf{x})$, a vector of size $s_p(d)$, and $\mathbf{v}_d(\mathbf{x})^TM_d(\mu_n)^{-1}\mathbf{v}_d(\mathbf{x})$. This being, this implies that the complexity and memory size growths are essentially exponential with $d$ or $p$, limiting the application to low dimensions and low degrees.

\begin{figure*}[!htbp]
\begin{subfigure}{0.45\textwidth}
    \includegraphics[width=\linewidth]{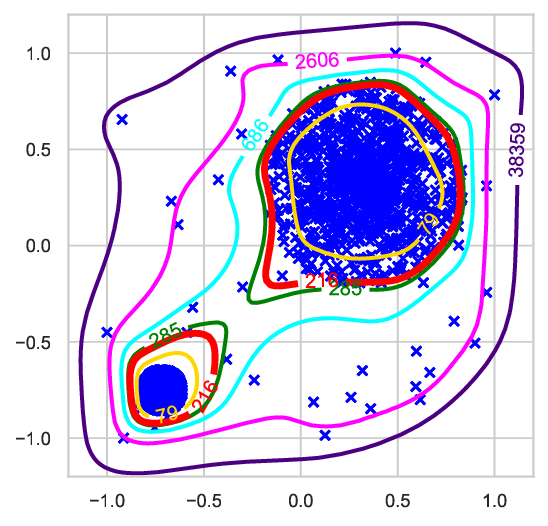}
    \caption{CF levelsets} \label{fig:1a}
  \end{subfigure}%
  \hspace*{\fill}   
  \begin{subfigure}{0.45\textwidth}
    \includegraphics[width=\linewidth]{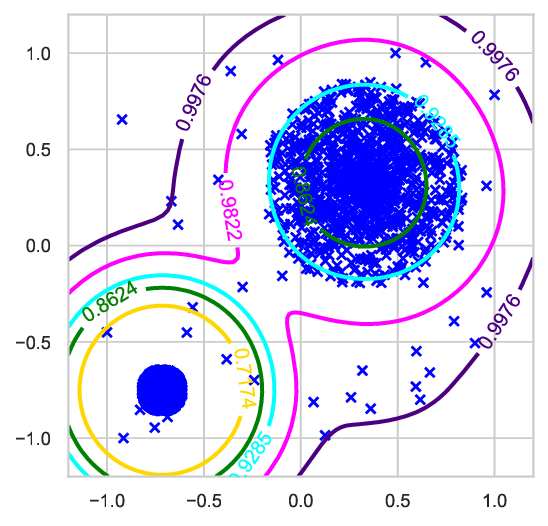}
    \caption{KDE levelsets} \label{fig:1b}
  \end{subfigure}
\caption{Level sets obtained for a dataset characterized by two disks and uneven densities with (a) CF using $d=6$ and (b) KDE (graph obtained using \textsc{Python} matplotlib library).}
\label{fig:cf_vs_kde}
\end{figure*}

\textbf{Illustrative example} -- In order to illustrate the behavior of the scoring function obtained, Figure \ref{fig:cf_vs_kde} compares scores from the CF with $d$=6 (Figure \ref{fig:cf_vs_kde}(a)) and scores obtained with KDE using the multivariate gaussian kernel (Figure \ref{fig:cf_vs_kde}(b)) on a dataset characterized by multiple densities. It consists of two clusters with different distributions; one is dense with 5000 samples circumscribed in a small circle and the other is sparse with 1000 samples circumscribed in a larger circle. On top of that, 50 points acting like outliers are sampled from a uniform distribution with its support around the two disks.

Fig.~\ref{fig:cf_vs_kde} clearly shows firsthand that the level sets generated by CF smoothly surround the cloud of points and some nicely capture the two clusters. On the other hand, the level sets generated by KDE do not capture precisely the dense cluster. In addition, the level set that captures at best this cluster rejects entirely the sparse cluster.

For a more rational evaluation, Table~\ref{tab:cf_vs_kde} considers the metrics AUROC, i.e., sensitivity (True Positive rate) versus specificity (False Positive Rate), and AP (Average Precision) approximating AUPRC, i.e., precision versus recall, that are recommended by \cite{ruff2021unifying} (section VII.B) for evaluating classification methods globally, independently of their tuning. Both scores are higher for CF. The results hence reinforce what is suggested by visual inspection of Fig.~\ref{fig:cf_vs_kde}, i.e., that CF is better at capturing the support of the cloud of points for this multi-density dataset.

Finally, note that the red thicker level set of CF, which nicely capture the support of the measure, corresponds to $\Omega_{\gamma_{d,p}}$ with $\gamma_{d,p}= d^{3p/2}$ dictated by the CF theory. 

\begin{table}[htbp]
    \centering
    \begin{tabularx}{0.5\textwidth}{YYY}
        \toprule
        Method & AUROC & AP \\
        \midrule
        CF & 0.9644 & 0.7250 \\
        KDE & 0.9372 & 0.6042 \\
        \bottomrule
    \end{tabularx}    
    \caption{AUROC and AP results obtained for CF and KDE on the two disks dataset}\label{tab:cf_vs_kde}
\end{table}


\section{Adapting the Christoffel Function to data streams}\label{sec:datastreams}

\subsection{Fast Model Update}\label{sec:dycf}

Most of the peculiarities of data streams listed in Section~\ref{sec:related}, like \emph{transiency}, \emph{infinity}, \emph{arrival rate}, and \emph{embeddedness}, boil down to requiring an efficient low computation and low memory incremental method. 

From the computational point of view, interestingly the CF complexity does not depend on the number of points but is essentially exponential in the number of dimensions $p$ and the chosen degree $d$. CF is hence expected to be competitive in low dimensions and for relatively small degrees.

From the memory point of view, the infinity of data streams is accounted by the use of the moment matrix $M_d(\mu_n)$ which contains the statistics of all the points without need to keep them in memory.

The capital gain of DyCF is incrementality and the ability of dealing with \emph{concept drift}, i.e., to update the model so that it follows any change in the distribution. The moment matrix given by Equation~(\ref{eq:empirical_M_d}) can be rewritten with the incremental formula 

\begin{multline} \label{eq:inc_mmatrix}
M_d(\mu_{n+1})=\frac{1}{n+1}[nM_d(\mu_n)\\+\mathbf{v}_d(\mathbf{x}_{n+1})\mathbf{v}_d(\mathbf{x}_{n+1})^T]. 
\end{multline} 

The CF outlier score given by Equations~\eqref{eq:Q} and \eqref{eq:inv_cf} requires to invert the moment matrix. Interestingly, the Sherman-Morrison formula provides an incremental way to invert a matrix of the form \eqref{eq:inc_mmatrix} as follows 

\begin{equation} \label{eq:sherman}(A+uv^T)^{-1}=A^{-1}-\frac{A^{-1}uv^TA^{-1}}{1+v^TA^{-1}u} 
\end{equation} 

Considering $A=nM_d(\mu_{n})$ and $u=v=\mathbf{v}_d(\mathbf{x}_{n+1})$, this leads to

\begin{multline} ((n+1)M_d(\mu_{n+1}))^{-1} = (nM_d(\mu_{n}))^{-1} \\- \frac{(nM_d(\mu_{n}))^{-1} \mathbf{v}_d(\mathbf{x}_{n+1}) \mathbf{v}_d(\mathbf{x}_{n+1})^T (nM_d(\mu_{n}))^{-1}} {1+ \mathbf{v}_d(\mathbf{x}_{n+1})^T (nM_d(\mu_{n}))^{-1} \mathbf{v}_d(\mathbf{x}_{n+1})}\\\label{eq:rank_one_update} 
\end{multline}

Equation~\eqref{eq:rank_one_update} can be used to compute the inverse CF $\Lambda^\mu_d(\mathbf{x})^{-1}$ in an incremental way, defining the proposed Dynamic Christoffel Function method named DyCF. 

It is important to note that DyCF requires only one parameter to be chosen, which is the degree $d$. The theory then dictates to use the level set defined by $\Omega_{\gamma_{d,p}}$. The DyCF scoring function is hence defined as $\Lambda^\mu_d(\mathbf{x})^{-1}$ normalized by $\gamma_{d,p}$
\begin{equation}\label{eq:dycf_scoring}
S_{d,p}(\mathbf{x})=\frac{\Lambda^\mu_d(\mathbf{x})^{-1}}{\gamma_{d,p}}, 
\end{equation}

\noindent from which a point $\mathbf{x}$ is defined as an outlier if $S_{d,p}(\mathbf{x})\geq 1$.

\subsection{Tuning Free}

Tuning-free is a highly desirable property that can be considered the holy grail in machine learning. Yet, as far as we know, it is not achieved by any outlier detection method. Interestingly, the evolution of the CF score, obtained for different values of $d$, has been theoretically characterized. The proposed Dynamic Chistoffel Growth method, named DyCG, leverages this property to achieve an efficient tuning-free method.

For $\mathbf{x}\in\mathbb{R}^p$, fixed, the evolution of $\Lambda^\mu_d(\mathbf{x})^{-1}$ as $d$ increases depends critically on whether $\mathbf{x}$ is in the support of $\mu$ or not. More precisely, for every $\mathbf{x}\notin \Omega$, the function $\mathbf{x}\mapsto \Lambda^\mu_d(\mathbf{x})^{-1}$ grows \emph{exponentially fast} with $d$, while its growth is \emph{at most polynomial} for $\mathbf{x}\in\Omega$.

The distinguishing property of exponential growth with $d$ for $\mathbf{x}$ outside the support of the measure is quantified by Theorem \ref{theorem}.
\begin{theorem}\label{theorem}(\citep{lasserre2022christoffel} Lemma 4.3.1 p.50)
Let $\mu$ be a positive Borel measure supported on the compact set $\Omega\subset\mathbb{R}^p$, and let $\mathbf{x}\not\in \Omega$ and $\delta>0$ be such that $\mathrm{dist}(\mathbf{x},\Omega)>\delta$. Then
\[\Lambda^\mu_d(\mathbf{x})^{-1}\,\geq\,s_p(d)2^{\frac{\delta d}{\delta+\mathrm{diam}(\Omega)}-3}\,d^{-p}\,(\frac{p}{e})^{p}\,e^{-p^2/d}\,.\]
\end{theorem}

At the same time, the magnitude of the CF score for points inside the support is at most polynomial in $d$ for $p$ fixed according to Theorem \ref{theorem2}.
\begin{theorem}\label{theorem2}(\citep{lasserre2022christoffel} Lemma 4.3.2 p.51)
Let $\mu$ be a positive Borel measure supported on the compact set $\Omega\subset\mathbb{R}^p$, the closure of a bounded domain $U$ with nice boundary, and let $\mathbf{x}\in U$ and $\delta>0$ be such that $\mathrm{dist}(\mathbf{x},\partial U)\geq\delta$. Then
\[\Lambda^\mu_d(\mathbf{x})^{-1}\,\leq\,s_p(d)\frac{C_p}{\delta^p}(1+p)^3\,,\]
where $C_p$ does not depend on $d$ but only on $p$.
\end{theorem}

Based on the asymptotic results of Theorem~\ref{theorem} and Theorem~\ref{theorem2}, DyCG is designed to assess the outlierness of a point based on two DyCF models of degrees $d_{min}$ and $d_{max}$. $d_{min}$ is naturally taken equal to $2$ and $d_{max}$ is taken equal to $6$ to make the problem tractable and can be reduced according to the available memory. The score $S_{d,p}(\mathbf{x})$ defined in Section~\ref{sec:dycf} is used for both models. This way, if $\Lambda^\mu_d(\mathbf{x})^{-1}$ follows a growth in $d^{3p/2}$ at least, then $S_{d_{max},p}(\mathbf{x})\geq S_{d_{min},p}(\mathbf{x})$. The DyCG scoring function is hence defined as

\begin{equation}\label{eq:dycg_score}
S'_{d_{max},d_{min},p}(\mathbf{x})=\frac{S_{d_{max},p}(\mathbf{x})-S_{d_{min},p}(\mathbf{x})}{d_{max}-d_{min}}\,,\,    
\end{equation}
and a sample $\mathbf{x}$ is considered outlying if $S'_{d_{max},d_{min},p}(\mathbf{x})\geq0$.

Note that DyCG requires to maintain two DyCF models simultaneously, which leads to an increase in memory use. Nevertheless, because DyCG is based on the evolution of the score, the value of the degrees $d_{min}$ and $d_{max}$ of the two models can be fixed once and for all, making of DyCG a tuning-free method. 


\section{Evaluation}\label{sec:eval}

\subsection{Process description}

To assess the effectiveness of the two proposed methods, an evaluation procedure is delineated in this section. This evaluation involves examining two types of data streams: synthetic data streams with labeled data, and real-world data streams without labels. All data streams can be found in the Git repository featuring our experiments\footnote{Code available on github~\citep{ducharlet2023odds}.}.

\subsubsection{Synthetic data streams}

Using Markov chain logic, synthetic data streams simulating multi-modal behaviors are constructed. Modes are specified in a configuration file, with parameters indicating whether they follow a normal or uniform distribution. Transitions between modes are then defined, with assigned probabilities and shapes (logarithmic, linear, exponential). Outliers are generated using a similar process. Two types of outliers are considered:
\begin{itemize}
    \item Type-I outliers are random values uniformly distributed around normal behaviors with a specified occurrence probability;
    \item Type-II outliers are defined as a short offset from normal behavior with appearing and lasting probabilities, enabling their persistence across successive measurements.
\end{itemize}

\begin{figure}[htbp]
    \centering    \includegraphics[width=0.45\textwidth]{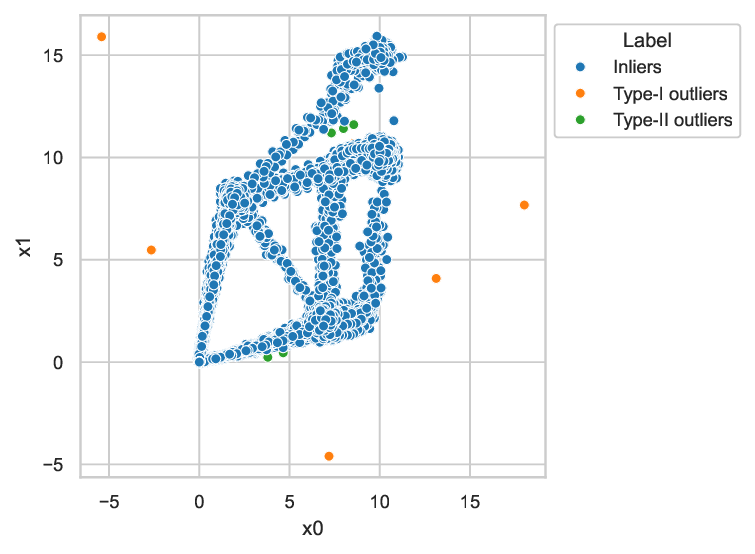}
    \caption{Samples issued from the first synthetic data stream setup. Blue dots represent normal behavior, orange dots are type-I outliers, green dots are type-II outliers.}
    \label{fig:synth_1}
\end{figure}

\begin{figure}[htbp]
    \centering    \includegraphics[width=0.45\textwidth]{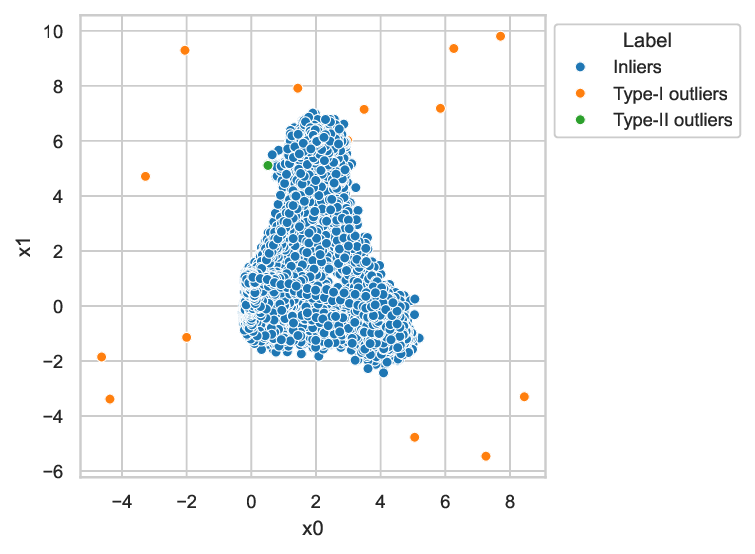}
    \caption{Samples issued from second synthetic data stream setup. Blue dots represent normal behavior, orange dots are type-I outliers, green dots are type-II outliers.}
    \label{fig:synth_2}
\end{figure}

Three setups are employed for generating synthetic data streams. Two of them are bivariate, showcasing samples illustrated in  Figures \ref{fig:synth_1} and \ref{fig:synth_2}, while the third is
trivariate in order to asses the effect of dimension on complexity. In each scenario, a behavioral alteration is introduced. This approach is intended to evaluate the model's capability to accommodate shifts from normal behavior. The alterations are as follows: 
\begin{itemize}
    \item in the first setup, a change in one mode's mean is implemented;
    \item in the second setup, an offset is applied to all data points;
    \item in the third setup, with three dimensions, a new mode is introduced at some point.
\end{itemize}

\subsubsection{Real-world data streams}

The real-world data streams originate from sensors installed on actual industrial luggage conveyor systems. The sensors specifically capture two physical variables: the speed of the conveyor belt and the intensity of the engine. 

\begin{figure}[htbp]
\centering
\includegraphics[width=0.4\textwidth]{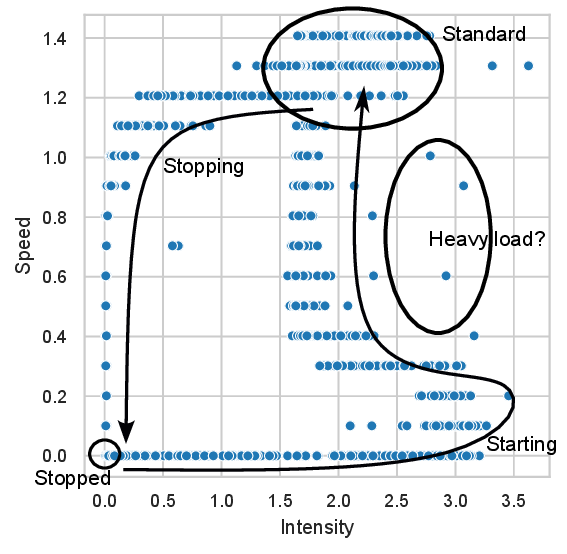}
\caption{Representation of a luggage\_conveyor data stream. Operating modes and transitions are visible. (graph obtained using \textsc{Python} matplotlib library and annotated manually)}
\label{fig:luggage_conveyor}
\end{figure}

These data streams exhibit distinct characteristics, consisting of three primary operational modes with nonlinear transitions between them (whose rationale guided the design of the afore mentioned synthetic datasets). The "stop" mode predominates, indicating the conveyor halted with both speed and intensity registering at zero. The "standard" mode reflects typical conveyor operation with nominal speed and intensity. An infrequent "heavy\_load" mode is also discernible, characterized by reduced speed and increased intensity to accommodate heavy luggage. Furthermore, transitions occur between the three operational modes, such as an intensity peak followed by a speed increase at the conveyor's start, and a fast decrease in intensity compared to speed when the conveyor stops. Visual representations of the data acquired for the various modes are provided in Figure~\ref{fig:luggage_conveyor}.

Five conveyors are considered with similar behaviors. All of them are working for seven successive days, with measurements issued every second (86400 samples per day). However, the data is sourced from wireless sensor networks and transmitted via radio transmissions with an unstable transmission frequency and potential packet losses. In this case, it is not critical as the exact measurement date is not considered (only the order of measurements is used). The packet loss rates during the measurement periods used for the five conveyors are respectively 11\%, 2\%, 3\%, 5\%, and 3\%.

\subsubsection{Evaluation process}

\begin{figure*}[htbp]
\centering
\includegraphics[width=0.95\textwidth]{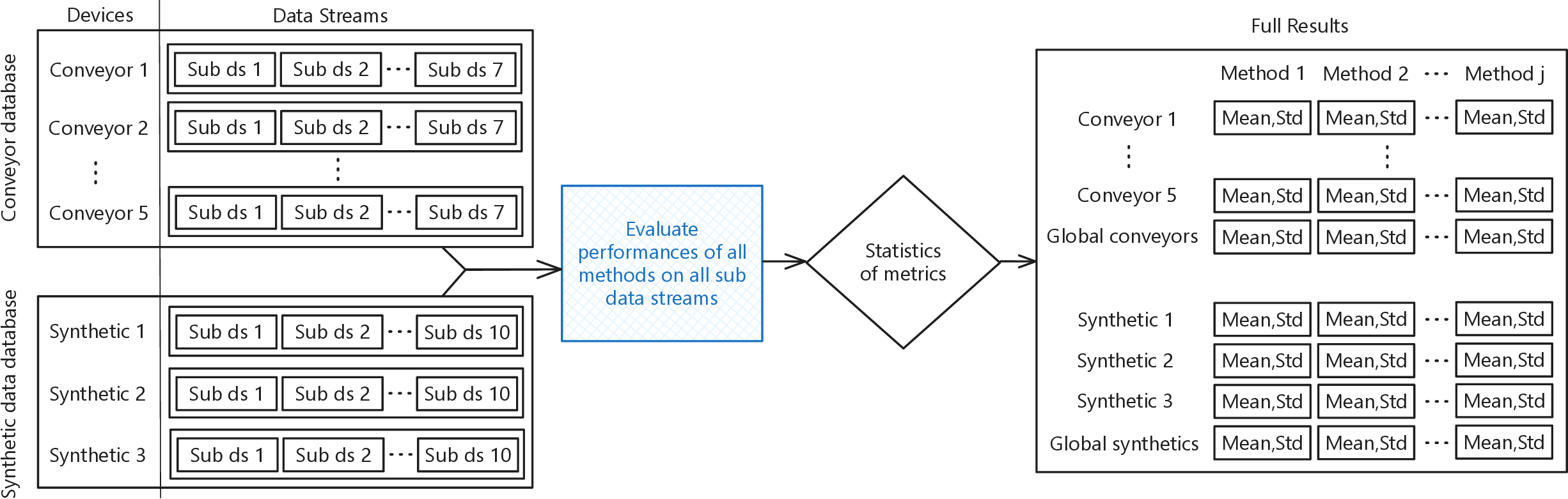}
\caption{Graph representing the full evaluation process.}
\label{fig:evaluation_process}
\end{figure*}

The whole evaluation process is described in Figure~\ref{fig:evaluation_process}. Data streams are organized in sub-data streams issued from different sources (different setups for the synthetic data streams and different conveyors for the real-world ones). 

Synthetic data streams are composed of 200k points divided in 10 sub-data streams while conveyor data streams are each divided in 7 working days. 

\begin{figure*}[htbp]
\centering
\includegraphics[width=0.9\textwidth]{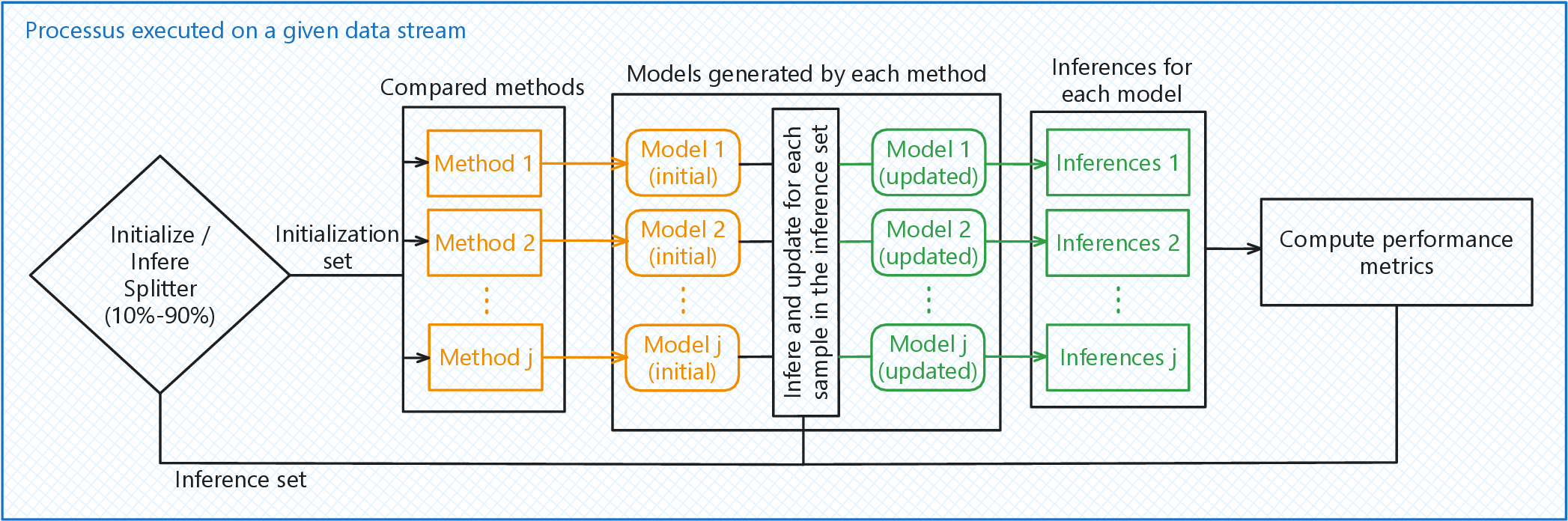}
\caption{Graph describing the process executed for each method.}
\label{fig:process_by_method}
\end{figure*}

\begin{figure}[htbp]
\centering
\includegraphics[width=.4\textwidth]{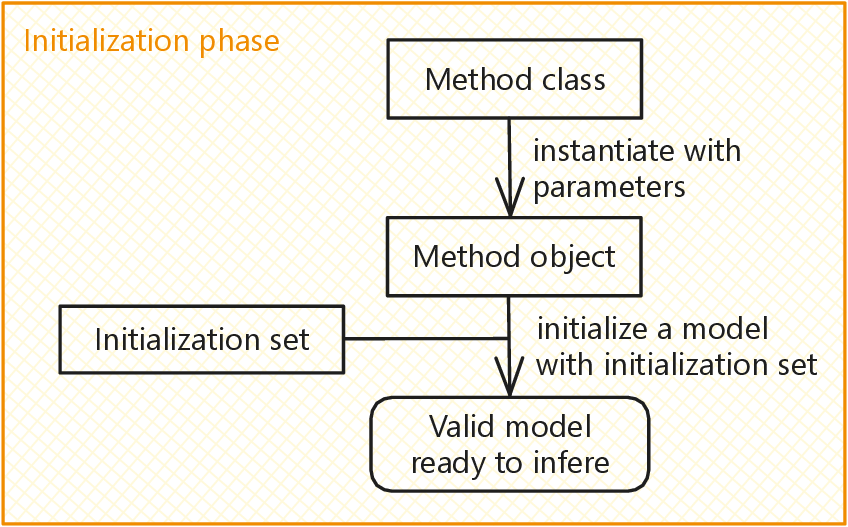}
\caption{Graph describing the initialization phase of a model.}
\label{fig:initialization_phase}
\end{figure}

\begin{figure}[htbp]
\centering
\includegraphics[width=.38\textwidth]{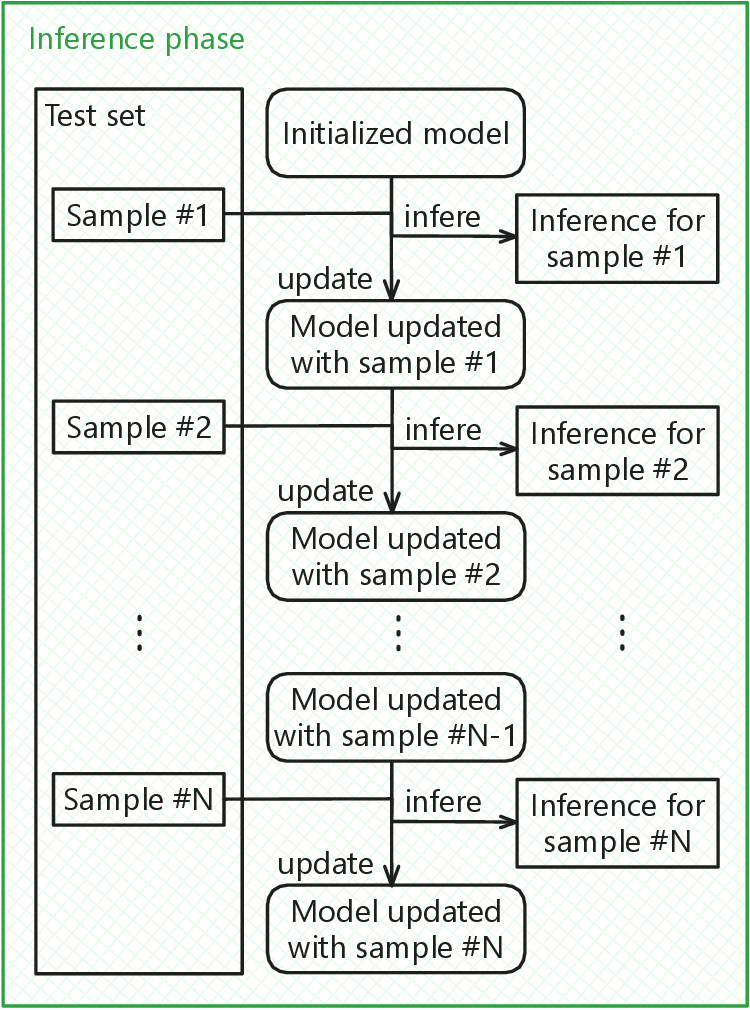}
\caption{Graph describing the inference phase of a model.}
\label{fig:inference_phase}
\end{figure}

The process used to evaluate the performance of the methods on all sub-data streams is described in Figure~\ref{fig:process_by_method}. Sub-data streams are divided in an initialization set used to initialize models and an inference set used for evaluation. The initialization phase is described in Figure~\ref{fig:initialization_phase} while the inference phase is described in Figure~\ref{fig:inference_phase}.

Mean and standard deviation of all metrics are computed for each method and each data stream. 

\subsubsection{Evaluation metrics}

Different metrics are used depending on the availability of labels:
\begin{itemize}
    \item in the synthetic cases, labels are available and it is possible to use popular metrics such as AUROC and AP, already used in the illustration example of Section~\ref{sec:cf_scoring},
    \item for conveyor cases, where no labels are available, AUROC and AP cannot be used. Instead, unsupervised metrics EM and MV \cite{goix2016how}, using the Excess-Mass and Mass-Volume curves respectively, are used.
\end{itemize}

No considered metric is threshold-sensitive, meaning that the choice of the threshold parameters does not impact the obtained score. Higher values of AUROC, AP, and EM are preferred, whereas lower values are sought for MV. It is important to note that EM and MV evaluate the extent to which a scoring function aligns with the statistical distribution of samples, which is not suitable for evaluating certain methods.

Finally, the average processing duration of a data point (computation of its outlierness score and model update) is computed to assess the speed of the methods, a characteristic highly esteemed in data stream contexts

\subsection{Competing methods}

The selected methods for comparison with DyCF and DyCG are all renowned outlier detection techniques for data streams. Each method has been re-implemented by us\footnote{Code available on github~\citep{ducharlet2023odds}.}, with the exception of SmartSifter, which relies on the \textsc{Python} implementation found in~\cite{k2021smartsifter}. Intensive parameter combinations have been tested to get the best out of the tuning for the comparison. The retained parameters are given in Table~\ref{tab:param_table}, outlining the number of parameters that need to be tuned for each method and pointing out the ease of use of DyCF and DyCG. Competing methods are commented below:
\begin{itemize}
    \item \textbf{Kernel Density Estimation (KDE)} has been presented in Section~\ref{sec:related} and illustrated in the example of Section \ref{sec:cf_scoring} and in Fig. \ref{fig:cf_vs_kde}. In order to be applicable to data streams, a sliding window of the last arriving points is used. This approach aims to mitigate time complexity and memory usage. Because the bandwidth parameter $\mathbf{H}$ is set from the Scott's rule of thumb of \citep{palpanas2003distributed,scott1992multivariate}, there are only two parameters to tune, which are the size of the sliding window $W$ (the number of points contained in the window) and the threshold on the score (or density estimate).   
    \item \textbf{SmartSifter} is selected in its parametric version as presented in \citep{yamanishi2004line} and briefly in Section~\ref{sec:related}. In our experiments, likelihood was used as a scoring function. The different parameters to be tuned are the threshold on the score, the number $k$ of gaussians in the GMM, a discounting parameter $r$ and a stability parameter $\alpha$.
    \item \textbf{Distance-based outliers using KDE (DBOKDE)} is derived from the kNN principle described in Section \ref{sec:related}. To reduce the complexity of counting the elements in a neighborhood, the number of neighbors is estimated using kernel density estimation. This method has been proposed in \cite{palpanas2003distributed}.
    \item \textbf{Incremental Local Outlier Factor (iLOF)} as presented in Section~\ref{sec:related}, is implemented with R*-Trees \citep{beckmann1990r} to reduce the kNN search complexity, as recommended in \cite{pokrajac2007incremental}.\footnote{Note that iLOF was improved in \citep{na2018dilof} with the Density summarizing Incremental LOF (DILOF) that reduces, in theory, the complexity while maintaining accuracy. Note that this is only true with really small windows or if the deletion part of iLOF, that makes the use of sliding windows possible, is abandoned. Otherwise, DILOF is heavier than iLOF because of the ``density summarizing part'' that is executed every $\frac{W}{4}$ observations, $W$ being the window length. For this reason, the comparison is done with iLOF solely.}
\end{itemize}

\begin{table}[!htbp]
    \centering
    \begin{tabular}{ccc}
        \toprule
        Method & Parameters & Values \\
        \midrule
        \multirow{4}{*}{KDE} & Threshold & Meaningless \\
         & Window size & 1000 \\
         & Kernel & Gaussian \\
         & Bandwidth & Scott's rule \\
        \midrule
        \multirow{4}{*}{SmartSifter} & Threshold & Meaningless \\
         & Nb components & 12 \\
         & Discounting parameter & 1e-3 \\
         & Stability parameter & 1.5 \\
        \midrule
        \multirow{5}{*}{DBOKDE} & Nb neighbors & Meaningless \\
         & Search radius & 0.1 \\        
         & Window size & 1000 \\
         & Kernel & Epanechnikov \\
         & Bandwidth & Scott's rule \\
        \midrule
        \multirow{7}{*}{ILOF} & Threshold & Meaningless \\
         & Nb neighbors & 10 \\        
         & Window size & 1000 \\
         & Min children & 3 \\
         & Max children & 12 \\
         & Reinsertion strategy & close \\
         & Reinsertion tolerance & 4 \\
        \midrule
        \multirow{2}{*}{DyCF} & Degree & 6 \\
         & C (threshold-like) & Meaningless \\
        \midrule
        DyCG & Degrees & (2, 6) \\
        \bottomrule
    \end{tabular}
    \caption{Table of parameters used in the experiments.}
    \label{tab:param_table}
\end{table}

\subsection{Results}

\subsubsection{Synthetic data streams}

\begin{figure*}[htbp]
    \centering    \includegraphics[width=0.95\textwidth]{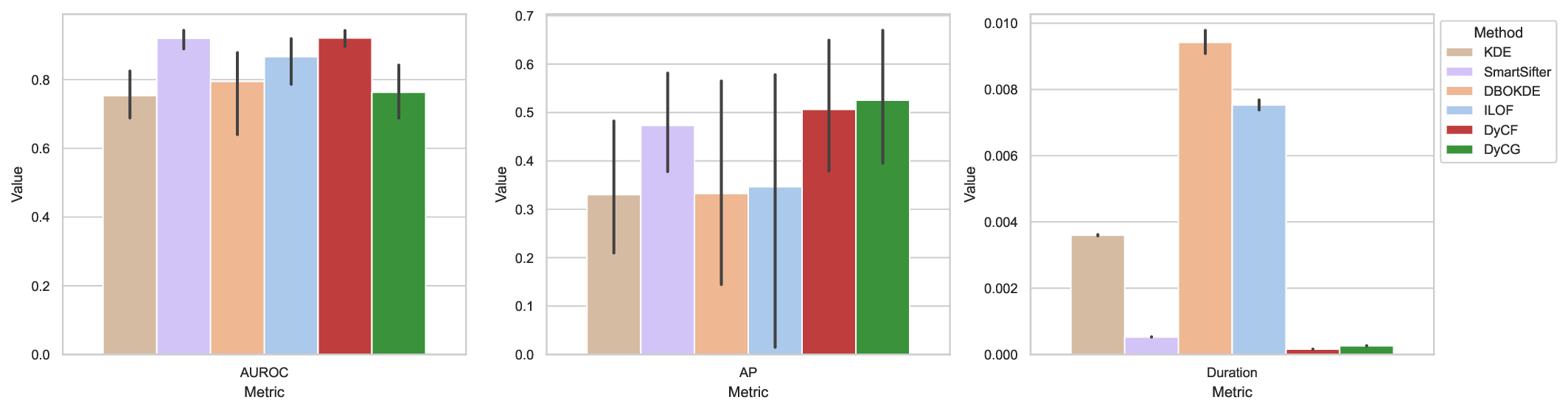}
    \caption{Results for the synthetic data stream first setup.}
    \label{fig:res_synth_1}
\end{figure*}

\begin{figure*}[htbp]
    \centering    \includegraphics[width=0.95\textwidth]{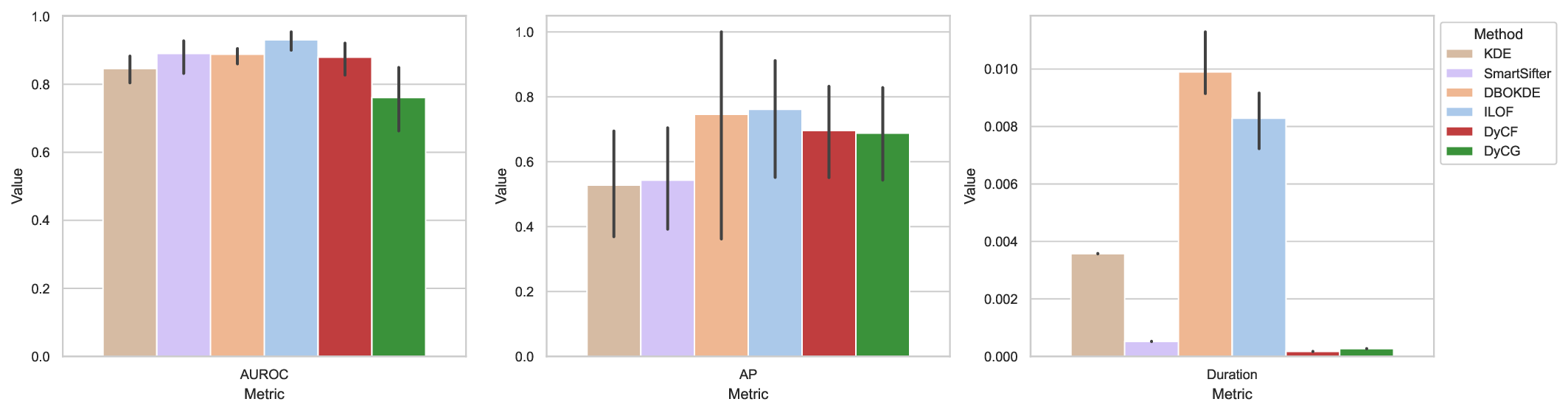}
    \caption{Results for the synthetic data stream second setup.}
    \label{fig:res_synth_2}
\end{figure*}

\begin{figure*}[htbp]
    \centering    \includegraphics[width=0.95\textwidth]{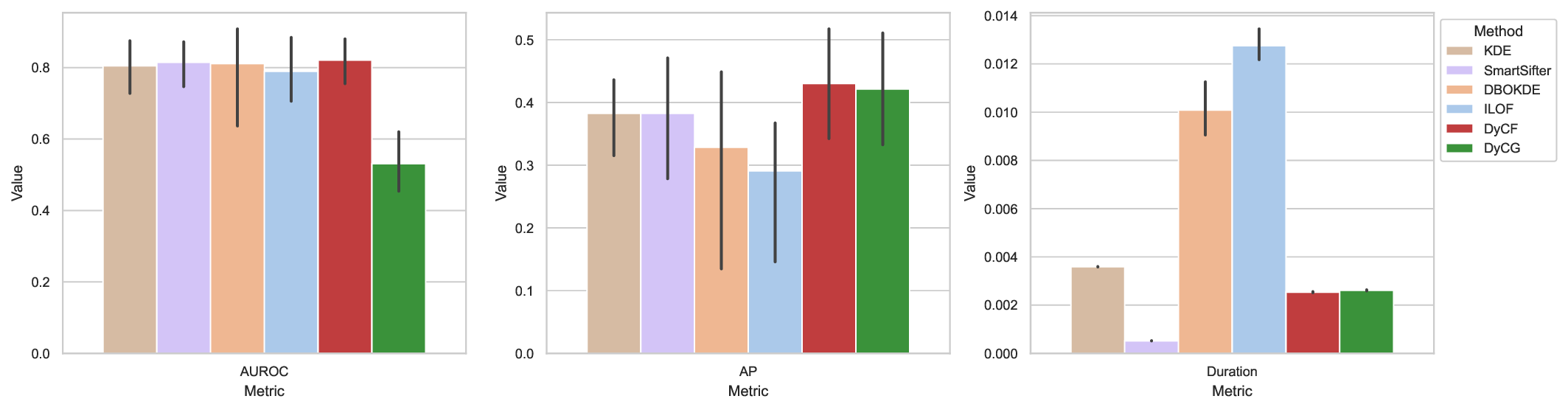}
    \caption{Results for the synthetic data stream third setup.}
    \label{fig:res_synth_3}
\end{figure*}

The results obtained for the synthetic data streams are shown in Figures~\ref{fig:res_synth_1}, \ref{fig:res_synth_2} and \ref{fig:res_synth_3} and values for each metric are given in Tables~\ref{tab:auroc_synth}, \ref{tab:ap_synth} and \ref{tab:time_synth}.

DyCF demonstrates performance at least on par (close to the best) with the compared methods concerning AUROC and AP metrics. Conversely, DyCG exhibits slightly lower performance on AUROC, particularly in the case of three-dimensional data streams; however, it yields superior results in terms of AP. 

Regarding the time metric, DyCF and DyCG outperform other methods when handling two-dimensional data streams, but SmartSifter is better with three-dimensional data streams. This is due to the dependence in $p$ of DyCF and DyCG.

\begin{table*}[!htb]
\centering
\scriptsize
\begin{tabular}{ccccccc}
\toprule
Dataset& KDE& SmartSifter& DBOKDE& ILOF& DyCF& DyCG\\
\midrule
Setup 1&  0.754 (0.116)&  0.920 (0.048)&  0.795 (0.136)&  0.867 (0.071)&  \textbf{0.921 (0.042)}&  0.763 (0.128)\\
Setup 2&  0.846 (0.070)&  0.890 (0.079)&  0.888 (0.025)&  \textbf{0.930 (0.028)}&  0.879 (0.077)&  0.761 (0.161)\\
Setup 3&  0.805 (0.124)&  0.815 (0.109)&  0.811 (0.151)&  0.790 (0.090)&  \textbf{0.821 (0.111)}&  0.531 (0.143)\\
\midrule
Global&  0.801 (0.109)&  \textbf{0.875 (0.092)}&  0.831 (0.111)&  0.862 (0.085)&  \textbf{\textit{0.874 (0.089)}}&  0.685 (0.178)\\
\bottomrule
\end{tabular}
\caption{AUROC mean (standard deviation in brackets) on synthetic data streams. Best value in bold and second best value in bold italic.}
\label{tab:auroc_synth}
\end{table*}

\begin{table*}[!htb]
\centering
\small
\begin{tabular}{ccccccc}
\toprule
Dataset& KDE& SmartSifter& DBOKDE& ILOF& DyCF& DyCG\\
\midrule
Setup 1&  0.330 (0.230)&  0.473 (0.174)&  0.333 (0.214)&  0.347 (0.295)&  0.507 (0.237)&  \textbf{0.525 (0.234)}\\
Setup 2&  0.528 (0.293)&  0.543 (0.272)&  0.746 (0.338)&  \textbf{0.761 (0.187)}&  0.696 (0.245)&  0.688 (0.253)\\
Setup 3&  0.383 (0.105)&  0.383 (0.161)&  0.329 (0.169)&  0.291 (0.125)&  \textbf{0.430 (0.154)}&  0.421 (0.154)\\
\midrule
Global&  0.413 (0.231)&  0.466 (0.211)&  0.469 (0.300)&  0.466 (0.290)&  \textbf{\textit{0.544 (0.237)}} &  \textbf{0.545 (0.238)}\\
\bottomrule
\end{tabular}
\caption{AP mean (standard deviation in brackets) on synthetic data streams. Best value in bold and second best value in bold italic.}
\label{tab:ap_synth}
\end{table*}

\begin{table*}[!htb]
\centering
\begin{tabular}{ccccccc}
\toprule
Dataset& KDE& SmartSifter& DBOKDE& ILOF& DyCF& DyCG\\
\midrule
Setup 1&  3.60e-3&  5.28e-4&  9.42e-3&  7.53e-3&  \textbf{1.60e-4}&  2.62e-4\\
Setup 2&  3.57e-3&  5.19e-4&  9.89e-3&  8.286e-3&  \textbf{1.71e-4}&  2.67e-4\\
Setup 3&  3.59e-3&  \textbf{5.21e-4}&  1.01e-2&  1.28e-2&  2.54e-3&  2.61e-3\\
\midrule
Global&  3.59e-3&  \textbf{5.23e-4}&  9.80e-3&  9.52e-3&  \textbf{\textit{9.56e-4}}&  1.05e-3\\
\bottomrule
\end{tabular}
\caption{Duration (in seconds per point) mean (standard deviation in brackets) on synthetic data streams. Best value in bold and second best value in bold italic.}
\label{tab:time_synth}
\end{table*}

\subsubsection{Real world data streams}

\begin{figure*}[htbp]
    \centering    \includegraphics[width=0.95\textwidth]{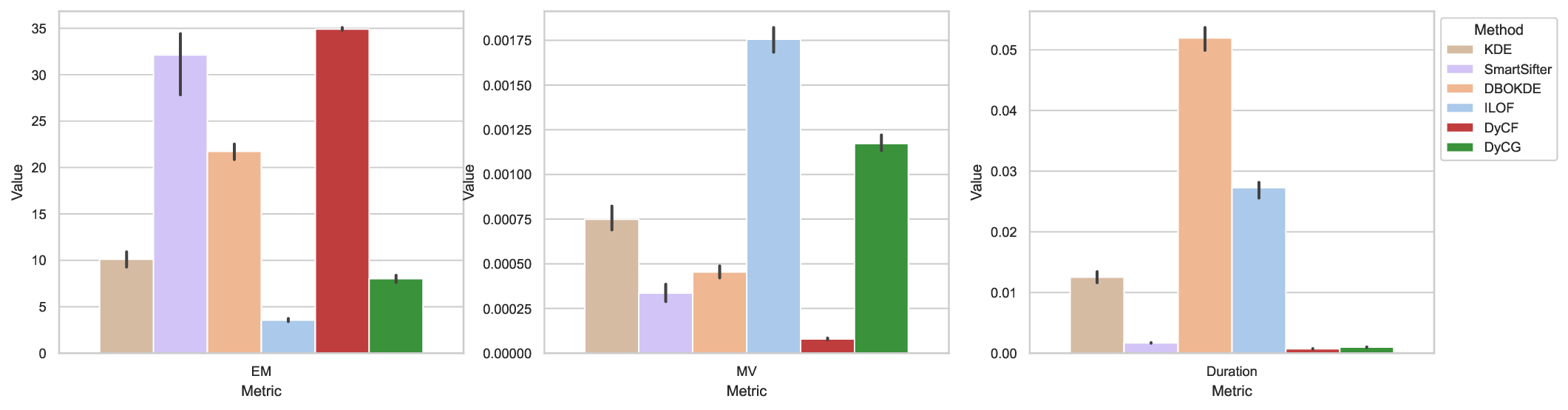}
    \caption{Results for the first conveyor data stream (provided by sensor node MOTE-47).}
    \label{fig:res_conv_1}
\end{figure*}

\begin{figure*}[htbp]
    \centering    \includegraphics[width=0.95\textwidth]{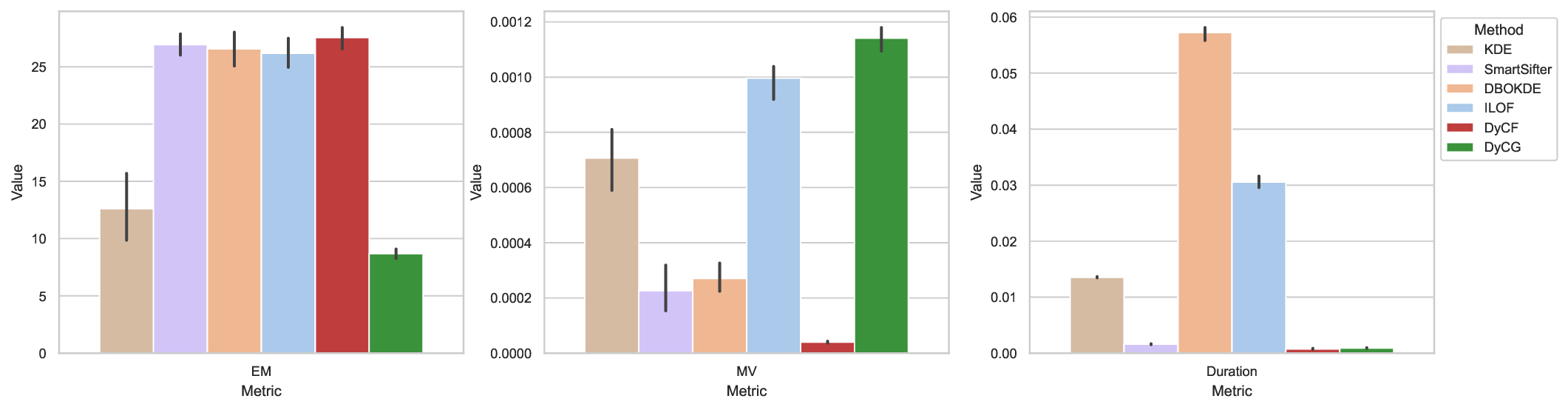}
    \caption{Results for the second conveyor data stream (provided by sensor node MOTE-67).}
    \label{fig:res_conv_2}
\end{figure*}

\begin{figure*}[htbp]
    \centering    \includegraphics[width=0.95\textwidth]{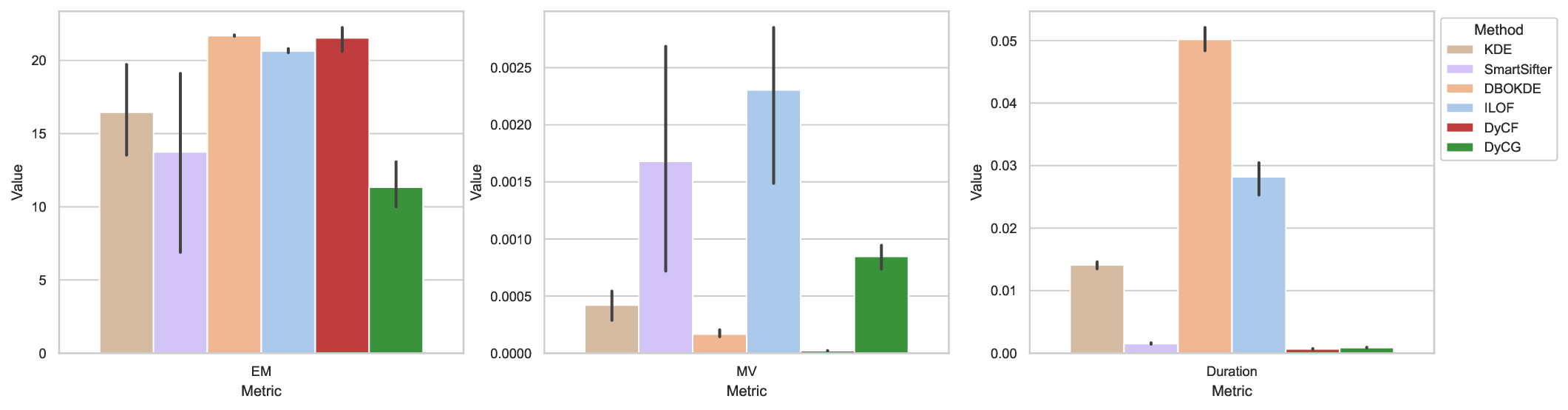}
    \caption{Results for the third conveyor data stream (provided by sensor node MOTE-72).}
    \label{fig:res_conv_3}
\end{figure*}

\begin{figure*}[htbp]
    \centering    \includegraphics[width=0.95\textwidth]{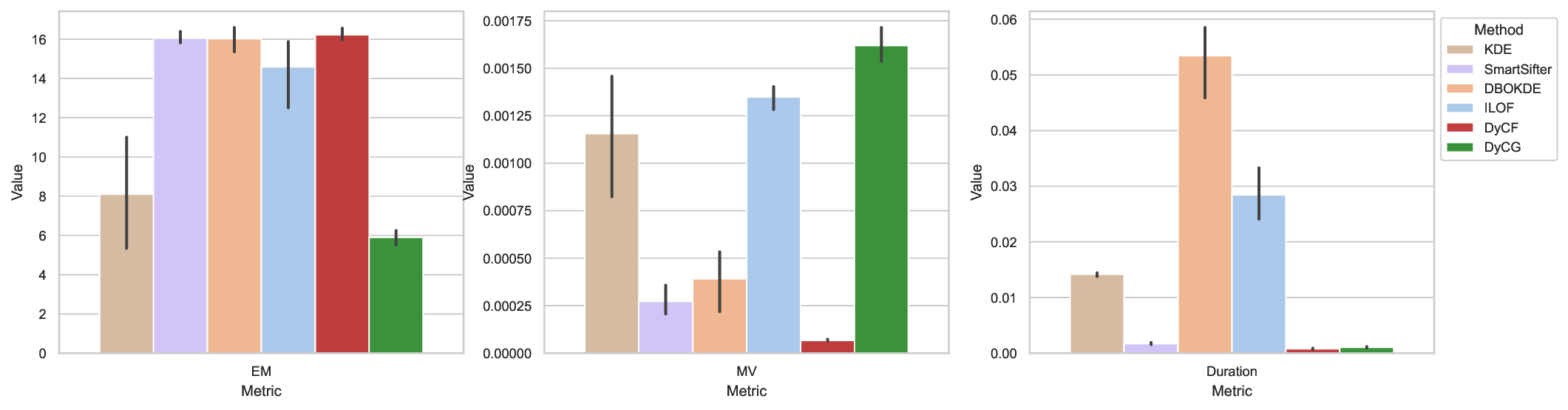}
    \caption{Results for the fourth conveyor data stream (provided by sensor node MOTE-75).}
    \label{fig:res_conv_4}
\end{figure*}

\begin{figure*}[htbp]
    \centering    \includegraphics[width=0.95\textwidth]{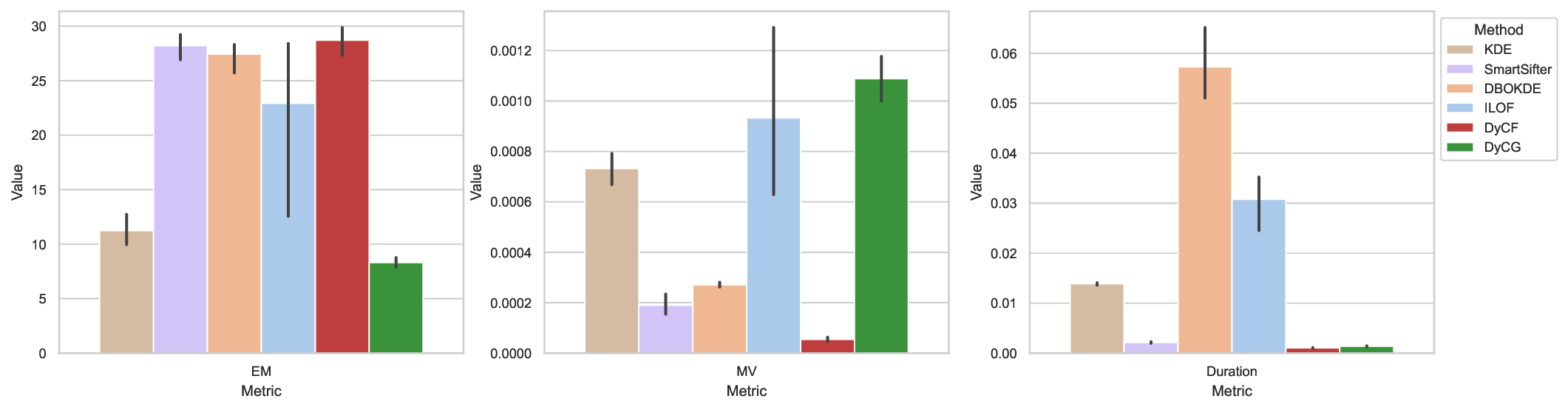}
    \caption{Results for the fifth conveyor data stream (provided by sensor node MOTE-78).}
    \label{fig:res_conv_5}
\end{figure*}

The results obtained for the conveyor data streams are shown in Figures~\ref{fig:res_conv_1}, \ref{fig:res_conv_2}, \ref{fig:res_conv_3}, \ref{fig:res_conv_4} and \ref{fig:res_conv_5} and values for each metric are given in Tables~\ref{tab:em_conv}, \ref{tab:mv_conv} and \ref{tab:time_conv}.

In this case study, DyCF provides by far the best results in all categories. On the other hand, DyCG offers significantly lower performances. The poor performance of DyCG can be explained by the underlying properties of EM and MV metrics. As a reminder, these metrics, designed for unsupervised anomaly detection, evaluate the alignment of the scoring function with the statistical distribution of samples, which is not in line with the transformation used to obtain DyCG's scoring function.

Interestingly, despite KDE's scoring being based on density estimation, KDE also exhibits poor performance on both EM and MV metrics. DBOKDE, which utilizes KDE at its core for estimating the number of neighbors, outperforms KDE on EM and MV.

\begin{table*}[!htb]
\centering
\small
\begin{tabular}{ccccccc}
\toprule
Dataset& KDE& SmartSifter& DBOKDE& ILOF& DyCF& DyCG\\
\midrule
Conveyor 1&  10.1 (1.234)&  32.1 (5.551)&  21.7 (0.136)&  3.56 (0.162)&  \textbf{34.9 (0.224)}&  8.02 (0.542)\\
Conveyor 2&  12.6 (4.292)&  26.9 (1.366)&  26.6 (1.477)&  26.2 (1.271)&  \textbf{27.5 (1.404)}&  8.68 (0.641)\\
Conveyor 3&  16.5 (4.455)&  13.8 (9.550)&  \textbf{21.7 (0.047)}&  20.6 (0.148)&  21.5 (1.184)&  11.3 (2.369)\\
 Conveyor 4& 8.11 (4.687)& 16.1 (0.458)& 16.0 (0.624)& 14.6 (1.824)& \textbf{16.2 (0.441)}&5.90 (0.586)\\
 Conveyor 5& 11.2 (2.056)& 28.2 (1.767)& 27.4 (1.489)& 22.9 (8.974)& \textbf{28.7 (1.932)}&8.31 (0.656)\\
 \midrule
Global&  11.7 (4.435)&  \textbf{\textit{23.4 (8.698)}}&  22.7 (4.328)&  17.6 (8.956)&  \textbf{25.8 (6.584)}&  8.45 (2.088)\\
\bottomrule
\end{tabular}
\caption{EM mean (standard deviation in brackets) on conveyor data streams. Best value in bold and second best value in bold italic.}
\label{tab:em_conv}
\end{table*}

\begin{table*}[!htb]
\centering
\begin{tabular}{ccccccc}
\toprule
Dataset& KDE& SmartSifter& DBOKDE& ILOF& DyCF& DyCG\\
\midrule
\multirow{2}{*}{Conveyor 1}&  7.49e-4 &  3.36e-4 &  4.54e-4 &  1.76e-3 &  \textbf{7.97e-5} &  1.17e-3 \\
 & (9.43e-5) & (7.23e-5) & (3.33e-5) & (6.93e-5) & \textbf{(7.23e-6)} & (6.06e-5) \\
\multirow{2}{*}{Conveyor 2} &  7.07e-4 &  2.26e-4 &  2.71e-4 &  9.96e-4 &  \textbf{3.99e-5} &  1.14e-3 \\
 & (1.657e-4) & (1.20e-4) & (5.17e-5) &(6.66e-5) & \textbf{(5.50e-6)} & (6.33e-5) \\
\multirow{2}{*}{Conveyor 3} &  4.20e-4 &  1.68e-3 &  1.67e-4 &  2.31e-3 &  \textbf{2.02e-5} &  8.47e-4 \\
 & (1.88e-4) & (1.47e-3) & (3.26e-5) & (8.19e-4) & \textbf{(3.56e-6)} & (1.54e-4) \\
 \multirow{2}{*}{Conveyor 4} & 1.16e-3 & 2.72e-4 & 3.91e-4 & 1.35e-3 & \textbf{6.74e-5 }&1.62e-3 \\
  & (4.64e-4) & (1.16e-4) & (1.60e-4) & (6.13e-5) & \textbf{(7.37e-6)} & (1.23e-4) \\
 \multirow{2}{*}{Conveyor 5} & 7.32e-4 & 1.91e-4 & 2.71e-4 & 9.33e-4 & \textbf{5.45e-5} &1.09e-3 \\
  & (9.07e-5) & (6.01e-5) & (9.55e-5) & (3.34e-4) & \textbf{(1.25e-5)} & (1.29e-4) \\
  \midrule
\multirow{2}{*}{Global} &  7.53e-4 &  5.41e-4 &  \textbf{\textit{3.11e-4}} &  1.47e-3 & \textbf{5.23e-5} &  1.17e-3 \\
  & (3.30e-4) & (8.52e-4) & \textbf{\textit{(1.23e-4)}} & (6.28e-4) & \textbf{(2.24e-5)} & (2.75e-4) \\
\bottomrule
\end{tabular}
\caption{MV mean (standard deviation in brackets) on conveyor data streams. Best value in bold and second best value in bold italic.}
\label{tab:mv_conv}
\end{table*}

\begin{table*}[htbp]
\centering
\begin{tabular}{ccccccc}
\toprule
Dataset& KDE& SmartSifter& DBOKDE& ILOF& DyCF& DyCG\\
\midrule
Conveyor 1&  1.25e-2&  1.69e-3&  5.20e-2&  2.73e-2&  \textbf{7.24e-4}&  9.96e-4\\
Conveyor 2&  1.35e-2&  1.60e-3&  5.72e-2&  3.05e-2&  \textbf{7.56e-4}&  9.20e-4\\
Conveyor 3&  1.41e-2&  1.54e-3&  5.02e-2&  2.82e-2&  \textbf{6.68e-4}&  8.76e-4\\
 Conveyor 4& 1.42e-2& 1.72e-3& 5.34e-2& 2.85e-2& \textbf{7.74e-4}&1.05e-3\\
 Conveyor 5& 1.39e-2& 2.15e-3& 5.73e-2& 3.08e-2& \textbf{1.06e-3}&1.39e-3\\
 \midrule
Global&  1.37e-2&  1.74e-3&  5.40e-2&  2.90e-2&  \textbf{7.97e-4}&  \textbf{\textit{1.05e-3}}\\
\bottomrule
\end{tabular}
\caption{Duration (in seconds per point) mean (standard deviation in brackets) on conveyor data streams. Best value in bold and second best value in bold italic.}
\label{tab:time_conv}
\end{table*}

\subsubsection{Discussion}

On the evaluated data streams, DyCF achieves state-of-the-art results while being easier to tune and faster than most methods.

DyCG allows to make tuning even easier but at great costs on performance. However, it gives the best results with AP. On top of that, its scoring function suffers from the underlying concept of EM and MV evaluations, and none of the employed metrics rewards the fact that DyCG does not require to set a threshold on the score, which is obviously a great advantage.

\begin{figure}[htbp]
    \centering
    \includegraphics[width=0.48\textwidth]{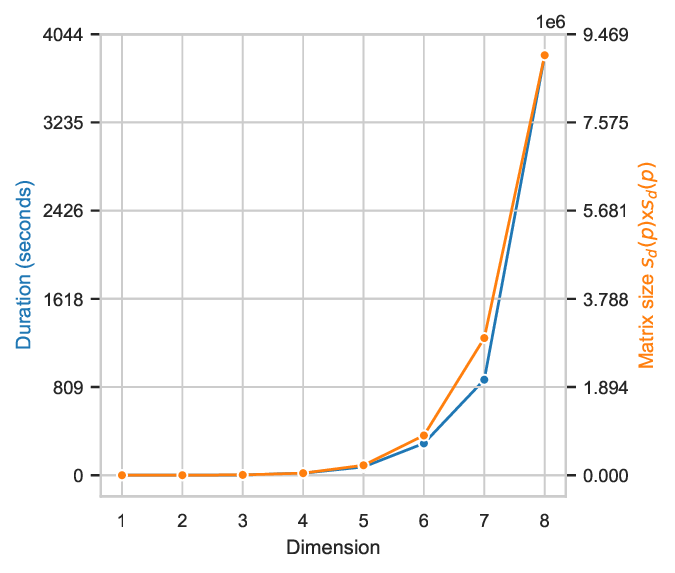}
    \caption{Growth of processing duration compared to matrix size when dimension $p$ increases}
    \label{fig:time-evolution}
\end{figure}

Regarding time complexity, the dependency in $p$ is noticeable between 2-dimensional datasets and the third synthetic setup which is 3-dimensional. To illustrate this further, in Figure~\ref{fig:time-evolution} we plot two graphs: (1) the processing duration by DyCF with $d=6$ of 500 data points drawn from uniform distributions of increasing dimensions $p$ and (2) the size of the moments matrix, which is $s_d(p)\times s_d(p)$ as a function of $p$.

\section{Conclusion and future work}\label{sec:conclu}

The principles on which methods discriminate normal points from outliers are paramount since they condition robustness and bias. 

In this article, two methods for unsupervised outlier detection in low dimensional data streams are proposed. Both leverage the properties of the Christoffel function and are built on solid theoretical foundations. 

The first method, DyCF, only requires two parameters to be tuned, while the second, DyCG, is completely free of tuning requirements. In this sense, the two methods elegantly remove the painful step of tuning, which is all the more painful in the unsupervised case and for typical non-stationary distributions of data streams. DyCF and DyCG have also shown great execution time and memory use performances, which has been noted of paramount importance. DyCF surpasses most of the methods it has been compared to, utilizing both well-established supervised metrics and lesser-known unsupervised ones.

These promising results encourage us to continue the work to overcome two limitations that were identified during this study: 
\begin{itemize}
    \item \emph{A numerical instability issue} has been observed for high values of $d$, i.e. for high degrees of the monomials that index the moment matrix. Actually, when the moment matrix has very small eigenvalues, some numerical instability occurs for its inversion. Future work will approach the problem in different ways and assess the impact on the accuracy of the resulting models:
    \begin{itemize}
      \item slightly perturbing the moment matrix by adding the identity matrix times a factor that makes the order of the resulting smallest eigenvalue reasonable for numerical inversion. This is known to bring more numerical stability as proposed in \citep{marx2021semi} (Eq. (8), p. 401) under the name of ``Tychonov regularization" . 
      \item replacing monomials by other polynomial basis. The use of Chebyshev polynomials of first kind would, in theory, give more numerical stability to the moment matrix. Typically, in the basis of monomials, the univariate moment matrix is Hankel and its multivariate analogue has a Hankel-like structure. Therefore for numerical computation, this choice of basis is not recommended in general, especially if the dimension of the matrix is large, in which case its inversion is severely ill-conditioned. Using the basis of Chebyshev polynomials is definitely better, as advocated (for many other purposes as well) in Chebfun \citep{driscoll2014chebfun}.
    \end{itemize}
    The numerical instability issue has been observed to be reinforced when using the Sherman-Morrison formula of Equation~\eqref{eq:sherman}, so for the evaluation section of this article we used the incrementation of the moment matrix and its inversion at each step. Theoretically, solving the numerical issue would mean being able to use the Sherman-Morrison formula, which would further reduce the time complexity of the two algorithms. Some experiments will be made in this direction.
    \item \emph{A scaling up issue} stemming from the size of the moment matrix, which is $\binom{p+d}{d}\times\binom{p+d}{d}$, where $p$ is just the problem dimension. This is why DyCF and DyCG are devoted to low dimensional outlier detection problems, as showcased by Figure~\ref{fig:time-evolution}. Nevertheless, the moment matrix size could be contained with a workaround consisting of randomly selecting a subsets of monomials. Future work will test this idea and assess its impact on the accuracy of the resulting models.
\end{itemize}

\section*{Acknowledgments} 
This project is related to ANITI through the French “Investing for the Future – PIA3” program under the Grant agreement n°ANR-19-PI3A-0004. 
Jean B. Lasserre's research is also part of the DesCartes' program supported by the National Research Foundation, Prime Minister's Office, Singapore under its Campus for Research Excellence and Technological Enterprise (CREATE) program.

The authors have no competing interests to declare that are relevant to the content of this article.

\bibliography{bibliography}
\bibliographystyle{IEEEtranN}

\end{document}